# A Text Classification Model Combining Adversarial Training with Pre-trained Language Model and neural networks: A Case Study on Telecom Fraud Incident Texts

Liu Zhuoxian[1]    Shi Tuo[1,2*]    Hu Xiaofeng[1]

[1]People's Public Security University of China, Information Network Security College, Beijing 100038, China

[2] Beijing Police College, Department of Public Security Management, Beijing 100190, China

**Abstract:**    Front-line police officers often categorize all police call reported cases of Telecom Fraud into 14 subcategories to facilitate targeted prevention measures, such as precise public education. However, the associated data is characterized by its large volume, diverse information content, and variations in expression. Currently, there is a lack of efficient and accurate intelligent models to replace manual classification, which, while precise, is relatively inefficient. To address these challenges, this paper proposes a text classification model that combines adversarial training with Pre-trained Language Model and neural networks. The Linguistically-motivated Pre-trained Language Model model extracts three types of language features and then utilizes the Fast Gradient Method algorithm to perturb the generated embedding layer. Subsequently, the Bi-directional Long Short-Term Memory and Convolutional Neural Networks networks extract contextual syntactic information and local semantic information, respectively. The model achieved an 83.9% classification accuracy when trained on a portion of telecom fraud case data provided by the operational department. The model established in this paper has been deployed in the operational department, freeing up a significant amount of manpower and improving the department's efficiency in combating Telecom Fraud crimes. Furthermore, considering the universality of the model established in this paper, other application scenarios await further exploration.

---

* Corresponding Author: Shi Tuo (Email: stshi8808@sina.com)

# 1  Introduction

In recent years, with the rapid development of the information society, major changes have taken place in the crime structure. Traditional crimes have continued to decline, while new forms of cybercrimes, such as telecommunications network fraud, have emerged as the predominant forms of crime[1]. According to statistics, in 2023, law enforcement agencies uncovered 437,000 cases of Telecom Fraud[2]. The procuratorate prosecuted 51,000 individuals for electronic fraud crimes, representing year-on-year increases of 66.9%. Courts at all levels have imposed severe punishments for domestic and overseas Telecom Fraud crimes, concluding 31,000 cases involving 64,000 individuals, a year-on-year increase of 48.4%[3]. In order to mitigate the high incidence of Telecom Fraud crimes, public security authorities have intensified their efforts to combat such crimes[4]. They have also improved the quality and efficiency of these efforts through various means. One such method is the pre-classification of police call reported incident texts, which categorizes Telecom Fraud crimes into 14 specific types. This classification enables separate filing and investigation of different types of cases, as well as the implementation of various measures to combat and prevent them. This approach has been widely promoted and recognized by public security authorities across the country and has played a positive role in practical operations.

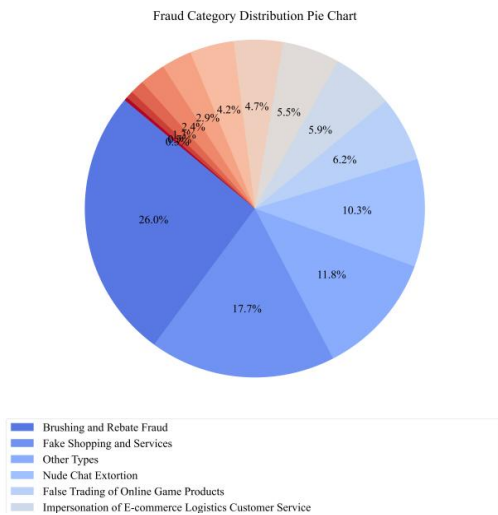

Fig. 1  Classification of 14 types of online fraud cases.

However, operational departments face numerous difficulties in processing police call reported incident data, especially Telecom Fraud crime data, due to limited technological means. Police call reported incident data originates from frontline alarm records and police feedback records, possessing a certain degree of real-time nature, with a large volume and rapid updates. Moreover, in practical work, due to public security organs encountering different subjects and situations during alarm reception, the recorded content has a certain degree of uncertainty, and the amount of information that can be obtained from the records also varies greatly. Furthermore, this type of text record is influenced by the subjectivity of the recording personnel, with different recorders having different textual expressions and styles. The existence of these practical situations results in a low degree of standardization of police call reported incident data, with the text being relatively free-form, posing significant challenges to the extraction of text features.

The text describing incidents of Telecom Fraud crimes represents a specific category of police incident data. While it shares similarities with general police incident data, it also presents unique challenges in processing. For example, due to the numerous subcategories within Telecom Fraud crimes, a large number of labels are required for classification. When data volume is limited, finding a suitable multi-class classification model becomes challenging. Additionally, among the 14 categories of Telecom Fraud crime incident data, many categories exhibit overlapping characteristics, with high similarity in textual descriptions between different categories. This similarity can lead to potential confusion during classification, significantly complicating the work of operational departments. As a result, law enforcement agencies have not yet identified an automated method to classify Telecom Fraud crime incident text. In practice, detailed classification of such text still relies on manual labeling of categories. At the grassroots level of public security agencies, a large number of police officers are required daily to perform this manual labeling, which is inefficient and consumes substantial law enforcement resources. Moreover, this manual labeling approach relies on the experience and cognition of individual officers, often leading to judgment errors.

In recent years, the rise of deep learning and artificial intelligence has provided a new approach to solving such problems. The earliest word vector models had certain limitations in handling complex text problems[5]. A series of text preprocessing algorithms, mainly based on BERT[6], have shown more outstanding results in text feature extraction. With continuous improvement and fine-tuning by researchers, more efficient BERT-like models have emerged, such as RoBERTa[7], MacBERT[8], and LERT[9], CNNs[10] (Convolutional Neural Networks) and RNNs[11] (Recurrent Neural Networks) are also used to extract different text features and have performed well in natural language processing tasks.

Therefore, this paper establishes a text classification model called LERT-CNN-BiLSTM. To further improve the model's robustness and generalization ability, we integrate the FGM adversarial training method commonly used in the image domain to perturb the embedding layer after preprocessing with LERT. The LERT model extracts three types of language feature information. In the text classification task of this paper, we extract the last layer output of LERT as the embedding. The embedding layer is input to the CNN and BiLSTM networks after adversarial training, extracting local features and contextual language sequence features of the text, respectively. Finally, the multi-channel fusion and parallel output are combined.

We conducted experiments on Telecom Fraud crime incident data in City B and achieved an accuracy of 83.9%, significantly outperforming existing typical text classification models. By training the model with a larger amount of data, we finally achieved an accuracy of 90% and deployed it in operational departments, freeing up a large amount of police resources. In addition, the text classification model proposed in this paper can also be applied to other text classification scenarios, providing a new approach and method.

In summary, this paper makes the following main contributions: First, it proposes a text classification model called LERT-CNN-BiLSTM. LERT extracts three types of language features to generate the embedding layer, and utilizes the parallel structure of CNN and BiLSTM to extract local features and contextual language sequence features, respectively. Second, it transfers the FGM adversarial training method from the image domain to perturb the embedding layer in natural language processing tasks, enhancing the robustness and generalization ability of the text classification model. Third, experiments using Telecom Fraud crime incident data in City B achieve a classification accuracy of 83.9%. Comparative ablation experiments demonstrate the significant superiority of the proposed model over other models, and it is ultimately deployed in operational departments. This work provides a new approach and method for other text classification tasks.

## 2 Related Work

Text classification is a typical task in natural language processing, where text (articles, comments, news, etc.) is automatically categorized into different classes according to predefined criteria[12-13]. It can address various issues such as sentiment analysis, question answering, and temporal prediction.

The earliest methods for text classification relied on statistical principles[14-15], probability theory[16], and rule matching[17]. However, traditional machine learning methods require manual feature extraction and cannot meet the needs of large-scale, long-sequence, and structurally complex text data classification scenarios[18]. In recent years, with the development of deep learning and artificial intelligence, more deep learning models and algorithms have emerged in the field of text classification[19]. We no longer need to manually design rules and features but instead use neural networks to extract semantic information and representations[13].

Generally speaking, current research on text classification tasks can be divided into the following directions:

Firstly, the application of pre-training techniques, followed by fine-tuning in downstream tasks, to enhance the model's language understanding capabilities and performance in specific tasks[20-22]. For example, popular pre-trained models such as BERT[6] and RoBERTa[7] not only automate feature extraction but also utilize Fine-tuning techniques[22] to adapt to different classification task scenarios.

Secondly, there is a continuous improvement in language models. In addition to traditional deep learning models such as recurrent neural networks[11,23], convolutional neural networks[10,24], attention mechanisms[25], and transformer models[26], the introduction and application of LLMs[27] (Large Language Models) in recent years have elevated the processing capabilities of language models for complex tasks and data to a higher level, becoming a core technology in the field of NLP (natural language processing). Examples include ChatGPT[28] model contributed by OpenAI and Google's LaMDA[29] model. However, due to their large parameter sizes and highly complex computational structures, they require significant computational power, posing challenges for practical deployment[30].

Thirdly, there is multimodal fusion[31], which leverages the knowledge of resource-rich modalities to assist in building models for resource-scarce modalities, thereby enhancing problem-solving capabilities[32-33]. However, this field not only relies on textual data but also requires a large amount of audio, image, and other types of data as training sets, which imposes certain requirements on the data used for training[34-35].

In this study, we need to classify police call reported incident text data, which is of significant importance for public security agencies in combating and preventing illegal crimes. As outlined in Section 1, we have described the challenges faced in handling police call reported incident data in practical scenarios, highlighting the urgent need for more technical solutions to address the current difficulties. However, these data are often tightly controlled by operational departments, resulting in a lack of research on police call reported incident data both domestically and internationally. In May 2020, Wang Mengxuan et al. improved the CRNN model and classified police incident texts, proposing a new method in this field[36]. However, further enhancement is needed in feature extraction. In May 2023, the Zhejiang Police College established the ERNIE-SA-DPCNN model for classifying online new type crime text data, achieving high accuracy[37]. However, its applicability is relatively limited and cannot meet the demands of various types of case classification. Our research fully leverages the models and viewpoints proposed by others but also introduces innovations to enhance the robustness and generalization capability of the model, making it suitable for various scenarios[38-39].

One approach is to learn from the commonly used adversarial training methods in the image domain[40-41]. Adversarial training was first applied in the image domain, and in recent years, some experts and scholars have transferred it to natural language processing tasks to enhance the robustness and generalization capability of models[42]. In the experimental process of text classification, we found that slight noise or interference in the input samples can lead to significant errors in the classification results. The adversarial learning approach generates adversarial samples at the embedding layer, aiming to minimize the loss function from a game-theoretic perspective. In 2017, Takeru Miyato first introduced this approach into natural language processing tasks[43], and later, popular adversarial learning methods such as FGM[44] and PGD[45] have also been gradually applied.

In the text classification task of this study, considering the limited computational resources in operational departments, it is challenging to support the deployment of large models. Additionally, research has confirmed that large models perform poorly in addressing the specific issues in this study. Furthermore, to ensure high text classification accuracy for better practical implementation and application, we based our approach on the well-preprocessed LERT model. We combined CNN and BiLSTM networks to extract local semantic information and contextual syntactic information, respectively, to enhance the model's feature extraction and classification capabilities. Moreover, to further improve the model's robustness and generalization capability for application in more scenarios, we introduced adversarial training. Specifically, we applied the FGM algorithm to perturb the embedding layer. Finally, through comparative experiments, we validated the reliability of our model.

## 3 Method

The diagram of the model architecture is shown below.

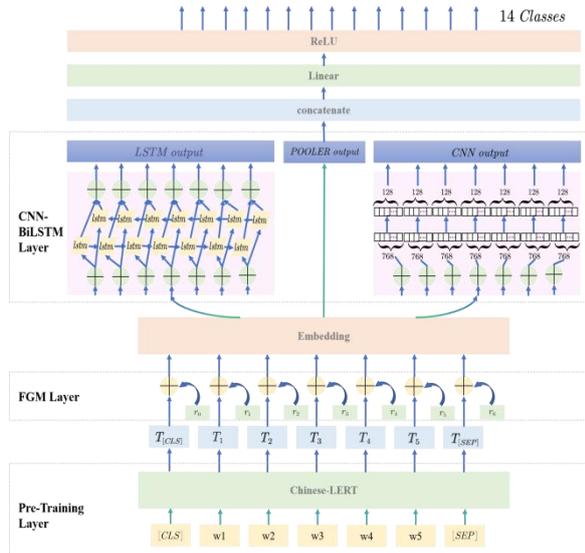

Fig. 2 Overall model architecture.

After preprocessing the Telecom Fraud police situation texts using the LERT model, the last layer is typically selected for the classification task. When generating the embedding layer, the FGM algorithm is introduced to perturb the output 768-dimensional word vectors. These vectors are then separately fed into the BiLSTM and CNN networks for training. The results from both networks are combined with the last layer output channel of LERT CONCAT. This approach extracts both contextual word order features and local text features. Finally, a fully connected layer is used for the 14-class text classification and output..

### 3.1 LERT Preprocessing

LERT[9] (Linguistically-motivated bidirectional Encoder Representation from Transformer) is a text preprocessing model developed by the joint efforts of Harbin Institute of Technology and University of Science and Technology of China in 2022. It aims to integrate multilingual features and improve upon the BERT-like text feature extraction model. By combining the WWM (Whole Word Masking) and NM (N-gram Masking) techniques, LERT further enhances the learning capability and downstream task performance of the masked language model.

LERT utilizes the LIP (Language Informed Pretraining) strategy to annotate the language labels of input text and employs three types of language features: POS (Part-of-Speech Tagging), NER (Named Entity Recognition), and DEP (Dependency Parsing) for text feature extraction. Additionally, it leverages the LTP (Language Technology Platform) to construct weakly supervised pretraining data. Following the extraction of language features, LERT undergoes multitask pretraining and simultaneously performs the original MLM (Masked Language Model) task.

The structure diagram of the LERT model is shown below:

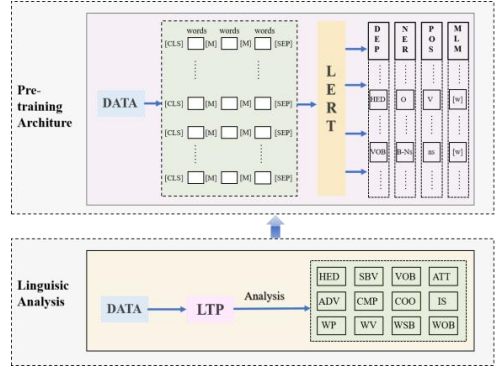

Fig. 3 LERT Model Architecture.

The LERT model first conducts linguistic analysis on the input alarm text data using the LTP (Language Technology Platform), a comprehensive natural language processing platform, to analyze syntactic relationships and linguistic features. In the second stage of the LERT model, it performs pretraining tasks, similar to most BERT-like models, by first tokenizing, adding CLS and SEP classification and sentence markers, and tokenizing. When generating masks, the LERT model combines two different masking techniques, WWM (Whole Word Masking) and NM (N-gram Masking), to better grasp the context of the text and understand the context more effectively. Unlike the BERT model, which only performs the MLM task during training, the LERT model also extracts three linguistic features: DEP (Dependency Parsing), NER (Named Entity Recognition), and POS (Part-of-Speech

Tagging). These features can perform syntactic analysis tasks, named entity recognition, and part-of-speech tagging tasks, respectively, enhancing the ability to understand text language. The multi-task training mode allocates training weights to each task through the LIP (Linguistically-Informed Pretraining) mechanism, ultimately determining the learning rate for each task, which conforms more closely to the general principles of human understanding and cognition. After performing the pretraining task, the LERT model outputs a fully connected layer and uses it as input for the subsequent classification model.

### 3.2 CNN

The CNN[46] (Convolutional Neural Networks) model extracts multi-layer features through convolutional layers and reduces dimensionality through pooling layers, exhibiting outstanding performance in classification problems and widely applied in computer vision, natural language processing, and other fields[47]. CNN learns using the traditional gradient descent method, training the input network with normalized and standardized data. The convolutional layers perform cross-correlation and linear convolution operations with multiple convolutional kernels to extract features of classification samples, while the pooling layers utilize pooling functions for feature selection and filtering[48]. The features are then nonlinearly combined through the fully connected layer to ultimately output the classification results. Traditional CNN models use the BP neural network as the basic framework for supervised learning. With the development of CNN in recent years, many improved CNN deep learning model frameworks have emerged[49]. CNN is characterized by its ease of training and optimization, as well as its strong robustness[50]. Many scholars at home and abroad use CNN for police text classification and have achieved good results, such as the CRNN police call text classification model improved by Wang Mengxuan[36] and others. In the CNN structure established in this paper, based on the features of text data, one-dimensional convolutional kernels are selected for feature extraction. The 768-dimensional vectors generated by the LERT model's embedding layer are used as the input to CNN, with 1 convolutional kernel for feature extraction, a sliding window of 1, and a total of 128 convolutional kernels. This CNN structure better extracts the local features of text, thereby enhancing the model's classification ability. For each 768-dimensional input sequence, it needs to be dot-multiplied with 128 convolutional kernels, and the output convolution results as the window slides. The formula for one-dimensional convolution is as follows:

$$Y_k[i] = \sum_{j=0}^{767} X[j] \cdot W_{k,j} \quad (1)$$

In equation(1), X[j] represents the element at position j of the input sequence X, $W_{k,j}$ denotes the weight of the k-th convolutional kernel at position j, Each element of the input sequence is multiplied by the corresponding weight of each convolutional kernel at the same position, and this operation is performed for each dimension of the 768-dimensional input sequence, followed by weighted summation. $Y_k[i]$ denotes the output of the k-th convolutional kernel at position i. Eventually, we obtain a 128-dimensional output sequence.

### 3.3 BiLSTM

LSTM[51](Long Short-Term Memory) is a new and efficient neural network sequence model based on gradient learning methods, which solves the long-term learning dependency problem of recurrent neural network algorithms. The LSTM algorithm introduces memory cells and gate control units to store historical information and long-term states, using gate control to control the flow of information[52]. The gates are divided into three types: input gate, forget gate, and output gate[53]. Gates are structures composed of nonlinear activation functions that control how much information is passed. Through gating, LSTM can achieve information storage and transmission. Generally, the operation process of LSTM can be divided into three stages: the first stage is the forget stage, where the forget gate of LSTM selectively discards some information, affecting the storage and transmission of information in the next memory cell. The second stage is the selective memory stage, where the information recording gate selectively remembers the input data and adds it to the information transmitted by the forget stage. The third stage is the output stage, where an activation function z scales the cell state, and the final output is controlled by another gate. Due to the superiority of LSTM in processing sequential data, it has been widely used in text information processing tasks. The structure of each LSTM unit is shown in the figure below.

In recent years, classical LSTM models have been continuously modified and transformed[54], and various variants such as LSTM models with peephole connections[55], Spatiotemporal LSTM[56], and bidirectional LSTM[57] models have emerged. Among them, the BiLSTM model achieves both forward and backward propagation. In text processing tasks, it considers both the information of the previous text and the information of the subsequent text, thereby better understanding the contextual semantics[57].

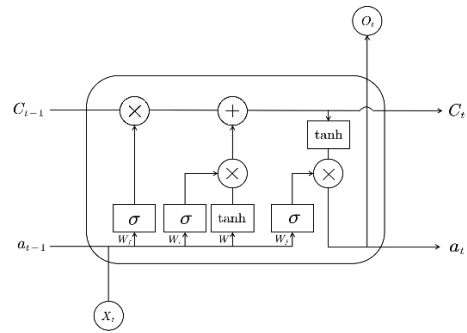

Fig. 4 LSTM Cell Structure.

The BiLSTM network established in this study is illustrated in Figure 4, where the input is a 768-dimensional embedding generated by the LERT layer, and the hidden layer is set to 64. Each vector is fed into the bidirectional propagation mechanism for training. In the network structure of Figure 4, $C_{t-1}$ represents the output of the previous state, which is the output at the previous time step t-1, and $X_t$ is the word vector of the current input at time step t. $a_{t-1}$ represents the hidden layer state at the previous time step. $C_t$, $O_t$, and at all represent the output generated after incorporating the input at the current time step. In the BiLSTM architecture, the embedding at each time step is fed into both the forward and backward LSTM sequences, and the outputs of both sequences are then concatenated, resulting in a 128-dimensional output vector.

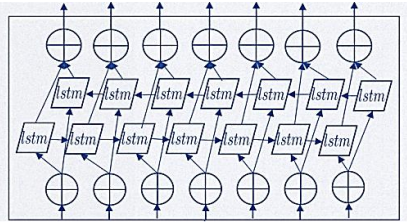

Fig. 5　BiLSTM network structure.

### 3.4　FGM Adversarial Training Method

The adversarial training method originated in the field of computer vision to address the issue of low accuracy in deep learning algorithms for classification[58]. It introduces noise, namely adversarial examples, to effectively regularize the models. In recent years, adversarial training has also been applied to natural language processing tasks to improve the robustness and generalization ability of models[43].

In text classification tasks, minor errors such as grammar and spelling mistakes can lead to significantly different classification results. This is particularly evident in the text data of telecommunications network fraud cases used in this study, where there is a large amount of similar word vector information and a wide range of subcategories, making confusion likely[39,59]. Applying adversarial training to generate adversarial examples can improve the classification ability of both original and adversarial examples[43]. While perturbations can be directly added to continuous image data, in natural language processing tasks, we perturb the entire embedding layer.

The principle of adversarial training can be summarized by the min-max formula[60], where perturbations are added in the direction of increasing loss while ensuring the overall minimization of the loss.

$$\min_{\theta} E_{(Z,y)\sim D}[\max_{|\delta|\leq \epsilon} L(f_{\theta}(X+\delta),y)] \quad (2)$$

In equation (2), $L(f_{\theta}(X+\delta))$ represents adding a perturbation $\delta$ to the sample $X$. After passing through the neural network function, the obtained output is compared with the label y to calculate the loss. The inner loss function is maximized to provide a perturbation that confuses the neural network more effectively. Once the inner-layer perturbation is determined, we minimize the entire loss in the outer layer and determine the values of the entire parameters.

The FGM (Fast Gradient Method)[44] algorithm is a typical adversarial training method, where the direction of perturbation ε is along the direction of the gradient to maximize the increase in loss. During normalization, the FGM method divides the value of each dimension of the gradient by the L2 norm of the gradient, strictly preserving the direction of the gradient. The calculation method for the perturbation in the FGM algorithm is shown below:

$$\delta = \epsilon \cdot (g/\|g\|_2) \quad (3)$$

Here, g represents the gradient of the loss function with respect to $X$, which is $\nabla_x L(f_{\theta}(X),y)$. Based on its structure, we can compute its gradient in network optimization. In this task, $X$ represents the input embedding matrix. We calculate the loss and the gradient g for the embedding layer generated by the LERT model, and then obtain the for this step by backward propagation. This $\delta$ is accumulated onto the embedding and used to compute the loss and backward propagation to obtain the adversarial gradient. However, since the current perturbation is optimal for the old parameters, we restore the embedding to its original value at this point without changing it. We then update the parameters during gradient descent. Through this approach, we believe that we have obtained the optimal perturbation, thereby minimizing the overall loss during adversarial training and enhancing the model's robustness and generalization ability.

## 4　Experiment and Discussion

### 4.1　Data

In this study, the 2023 Telecom Fraud crime data of a certain city were selected. The data were entered by grassroots public security organs according to the entry standards for grassroots telecommunication network fraud cases, and were statistically analyzed on a daily basis. After strict desensitization, the data were used for this study.

When entering Telecom Fraud cases, the data are usually divided into two parts. One part is the content recorded by the police call receiving officer, including the report time, victim information (gender, age, ID card number, caller's phone number, occupation, education level, employer, etc.), location of the case, type of fraud, amount of loss, etc. The other part is the feedback information obtained during the case acceptance through follow-up visits, which is more comprehensive and detailed. It may include the first contact platform with the suspect (network platform, SMS, incoming call), chat platform (chat account and nickname), brief description of the fraud script, transfer method (WeChat, Alipay, online banking, etc.), additional transfer information (specific number of transfers, amount, opponent information, and respective bank information), and other case-related information.

According to the classification standards of Telecom Fraud cases by public security organs, all cases are divided into 14 categories. The case data are shown in Table 1. For research purposes, this study used random sampling to extract 10,000 cases for statistical classification, ensuring that the sample and the population have the same data distribution. The case text was used as the classification text, and the case category was used as the classification label. The dataset was divided into training, validation, and test sets in an 8:1:1 ratio. The dataset was split using a random seed algorithm. When generating a random number sequence, the seed value 24 was used as the starting point to ensure that the generated random number sequence is the same every time, thus ensuring the repeatability and stability of the experiment results.

Table 1   Example of Police Case Text (take Fake Shopping and Services as an example)

| Summary of Police Cases |
| --- |
| On December 30, 2023, at 24:00, Mr. Changju in YD Town, MTG District, B City received a strange express delivery in Building 1, Unit 101 of a certain residential area, which contained a refund notice. Mr. Changju scanned the QQ code and downloaded the Zy agent APP for the refund. He was later defrauded of 20220.8 yuan in the APP refund. The actual loss was 20220.8 yuan. On December 31, 2023, at 29:52, Mr. Changju transferred 20020 yuan to the other party (Hlj Rural Credit Cooperative, Mr. Qinglong) through (China Ping An Bank, Mr. Changju, 000000000000000000). |

### 4.2  Environment and Settings

This study was developed using the Python programming language and the PyTorch deep learning framework. The transformers library was also used to load BERT and other preprocessing models. The LERT and other natural language preprocessing models have been open-sourced on GitHub and can be deployed locally. After multiple experiments, it was found that when the batch size is 4, the number of epochs is 3, and the learning rate is 1e-6, the model converges and achieves the optimal accuracy. The experimental settings are as follows:

Table 2 describes the environment configuration for model operation, Table 3 describes the parameter configuration of the CNN and BiLSTM models, Table 4 describes the parameter configuration of the LERT model.

Table 2   Environment and settings of the experiment

| Name | Version |
| --- | --- |
| numpy | 1.23.5 |
| openpyxl | 3.0.10 |
| pandas | 2.1.4 |
| python | 3.11.7 |
| scikit-learn | 1.3.0 |
| tokenizers | 0.15.0 |
| torch | 2.0.0+cu118 |
| transformers | 4.36.2 |
| operating system | Win11 |
| server | NVIDIA GeForce RTX 4060 Laptop |
| batch_size | 8 |
| learning rate | 1e-6 |
| epoch | 3 |
| optimizer | Adam |

Table 3   The parameter values of the CNN-BiLSTM model

| Name | Version |
| --- | --- |
| convolutional dimension | 1d |
| embedding layer dimension | 768 |
| number of convolutional kernels | 128 |
| kernel_size | 1 |
| stride | 1 |
| padding | 0 |

Table 4   The parameter values of the LERT model

| Name | Version |
| --- | --- |
| attention_probs_dropout_prob | 0.1 |
| classifier_dropout | null |
| directionality | bidi |
| hidden_act | gelu |
| hidden_dropout_prob | 0.1 |
| hidden_size | 768 |
| initializer_range | 0.02 |
| intermediate_size | 3072 |
| layer_norm_eps | 1e-12 |
| max_position_embeddings | 512 |
| num_attention_heads | 12 |
| num_hidden_layers | 12 |
| pad_token_id | 0 |

The experiments were evaluated using metrics such as Accuracy, Precision, Recall, F1 Score, and Average Loss on the Test Dataset. These metrics provide a comprehensive evaluation of a model's performance in classification tasks. Accuracy measures the overall correctness of predictions, while precision and recall focus more on the model's ability to distinguish between positive and negative samples. The F1 Score combines precision and recall to provide a balanced assessment of the model's performance. The principles and formulas for calculating

these metrics are as follows:

Accuracy refers to the proportion of correctly predicted samples out of the total number of samples, and its formula is:

$$Accuracy = \frac{TP+TN}{TP+TN+FP+FN} \quad (4)$$

Where TP (True Positive) is the number of true positive cases, TN (True Negative) is the number of true negative cases, FP (False Positive) is the number of false positive cases, and FN (False Negative) is the number of false negative cases.

Precision is the proportion of samples predicted as positive that are actually positive, and its formula is:

$$Precision = \frac{TP}{TP+FP} \quad (5)$$

Recall is the proportion of actual positive samples that are correctly predicted as positive by the model. It is calculated as:

$$Recall = \frac{TP}{TP+FN} \quad (6)$$

The F1 Score combines Precision and Recall into a single metric, which is their harmonic mean. It is calculated as:

$$F1\_Score = 2 \times \frac{Precision \times Recall}{Precision + Recall} \quad (7)$$

The average loss is the mean value of the loss function across all samples. It is typically used in regression tasks. In this paper, the cross-entropy loss function is used for error calculation. The principle of the cross-entropy loss function is as follows:

$$Loss = -\frac{1}{N}\sum_{i=1}^{N}(y_i \log(\hat{y}_i) + (1-y_i)\log(1-\hat{y}_i)) \quad (8)$$

Where N is the number of samples. $y_i$ is the true label of sample $i$. $\hat{y}_i$ is the predicted probability of sample being in the positive class. We calculate the average loss by averaging the loss values of all samples, which is referred to as Average Loss in this paper.

### 4.2 Results and discussion

This paper conducted multiple sets of comparative experiments to validate the various components of the model. In summary, the comparative experiments designed in this paper consist of the following parts: 1. Validation of the preprocessing part of LERT, which is superior to other BERT-like models in text feature extraction, enhancing the representational ability for classifying texts. 2. Ablation experiments to confirm the roles of the BiLSTM and CNN layers in feature extraction. 3. Comparative experiments with other typical lightweight models to demonstrate that the proposed model significantly outperforms existing typical text classification models. 4. Validation of the model's robustness and generalization ability by changing learning rates, batch sizes, and epochs, demonstrating the rationality and scientific nature of the parameter settings.

Firstly, experiments were conducted by replacing the LERT preprocessing part with different BERT-like models while keeping the rest of the model structure unchanged. The experimental results, as shown in Table 5, indicate that the LERT model performs better in the preprocessing task based on the five metrics. Where Accuracy Precision Recall F1 Score Average Loss is denoted by acc, P, R, F1, Loss as abbreviations respectively.

Table 5　Comparative Experiments on the LERT Preprocessing Part

| Name | Acc | P | R | F1 | Loss |
| --- | --- | --- | --- | --- | --- |
| **LERT-CNN-BILSTM** | **0.832** | **0.818** | **0.832** | **0.824** | **0.660** |
| MACBERT-CNN-BILSTM | 0.814 | 0.798 | 0.814 | 0.803 | 0.775 |
| ROBERTA-CNN-BILSTM | 0.822 | 0.810 | 0.822 | 0.813 | 0.696 |
| BERT-CNN-BILSTM | 0.826 | 0.813 | 0.826 | 0.818 | 0.713 |

Next, we conducted comparative experiments by removing the structures of the CNN and BiLSTM layers to confirm their roles in feature extraction. The experimental results, as shown in Table 6, indicate that both the CNN and BiLSTM layers play a role in feature extraction, with the CNN layer contributing more significantly to the improvement of the results. This suggests that, for police case text, local semantic features are more prominent than contextual syntactic features. This is consistent with our intuition, as semantic information is often more important in determining the category of a case.

Table 6　Ablation experiment

| Name | Acc | P | R | F1 | Loss |
| --- | --- | --- | --- | --- | --- |
| **LERT-CNN-BILSTM** | **0.832** | **0.818** | **0.832** | **0.824** | **0.660** |
| LERT-CNN | 0.821 | 0.815 | 0.821 | 0.814 | 0.667 |
| LERT-BILSTM | 0.721 | 0.668 | 0.721 | 0.688 | 0.927 |

On top of the baseline model, we fine-tuned the model by perturbing the embedding layer using the FGM adversarial training method, which generates adversarial samples. To verify the effectiveness of this method, we designed a control experiment. As shown in Table 7, the FGM method improved the model's accuracy across all four metrics.

Table 7　Control Experiment of FGM Fine-tuning Method

| Name | Acc | P | R | F1 | Loss |
| --- | --- | --- | --- | --- | --- |
| **LERT-CNN-BILSTM（+FGM）** | **0.839** | **0.834** | **0.839** | **0.835** | **0.592** |
| LERT-CNN-BILSTM | 0.832 | 0.818 | 0.832 | 0.824 | 0.660 |

In addition, we attempted to use lightweight classic text classification models to perform the task in this study. The experimental results showed that the lightweight models performed significantly worse than the model established in this paper for police case text classification tasks.

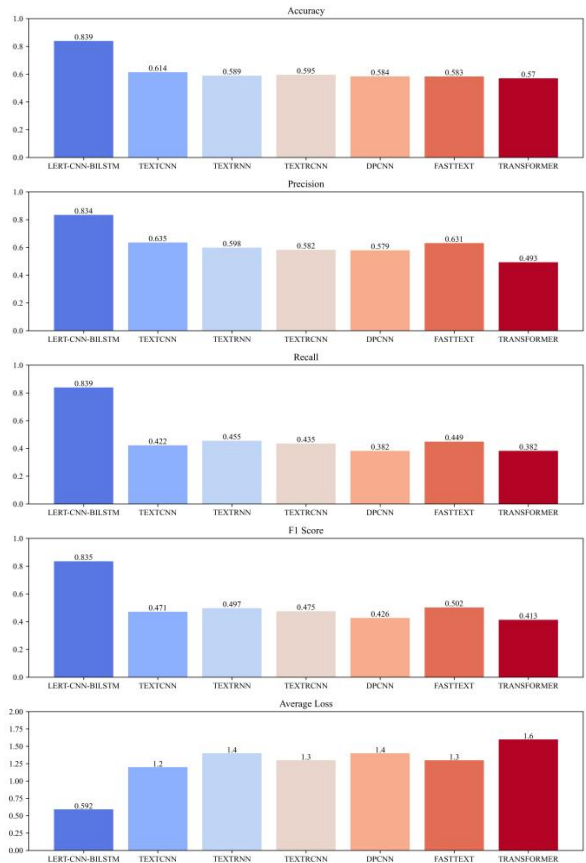

Fig. 6　Comparative Experiment Results of Typical Text Classification Models.

We tested with different learning rates and found that the model performed best when the learning rate was set to 1e-6.

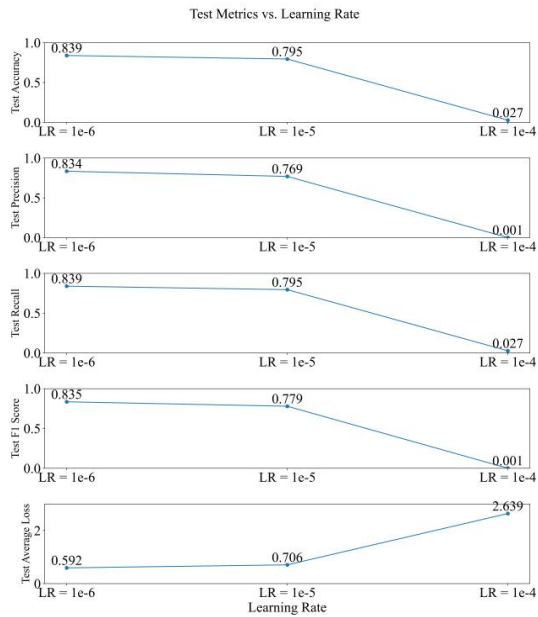

Fig. 7　Vary from the learning rate.

In addition to learning rate, batch size is also an important parameter in deep learning. Choosing an appropriate batch size is beneficial for ensuring the model's computational efficiency, convergence speed, and memory consumption. After consulting a large amount of literature, we initially set the batch size to 8 for the sake of convenience in the experiment. To validate the rationality and reliability of the parameters, we conducted experiments by setting the batch size to 4, 8, and 16 respectively, while keeping other settings unchanged. The results show that increasing or decreasing the batch size will lead to worse experimental results, confirming our initial hypothesis. We chose a batch size of 8 as the parameter for the model.

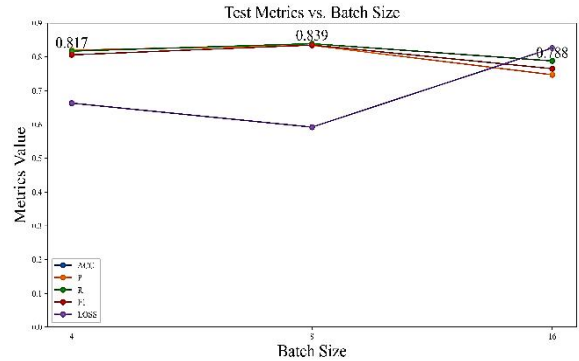

Fig. 8　Vary from the batch size.

In January 2024, based on the model architecture established in this paper, we trained the model using 150,000 police case text data from the practical department. The model parameters were saved, and an executable file was developed to automate the classification of input police case text using the model. The classification accuracy reached 90%, which largely meets the requirements of practical applications, saving a significant amount of police resources.

## 5　Conclusions

Telecom Fraud is one of the most prevalent types of crimes in recent years, and law enforcement agencies have actively employed various effective measures to combat it. One common approach is to classify the police case information received during emergency calls. Through this method, Telecom Fraud is subdivided into 14 categories, and targeted measures are taken based on the characteristics of each category to apprehend suspects. Simultaneously, preventive measures are taken to educate vulnerable groups, aiming to prevent such crimes before they occur. However, due to the non-uniformity and large volume of police case data, grassroots police officers still manually classify case types, lacking an efficient and practical automated method for case classification.

The model that combines the FGM adversarial training method with the LERT-CNN-BiLSTM architecture proposed in this paper demonstrate good classification performance in text data with high confusion, uncertainty, and varying lengths. By conducting comparative experiments using practical data, the model's excellence is reflected in all four metrics on the test set. With training on a large practical dataset, the model achieves an accuracy rate of over 90%, essentially meeting the requirements of practical applications and having certain practical significance. Additionally, the model established in this paper can be applied to other text classification scenarios, although further exploration and application are needed in other practices.

It should be noted that the algorithm in this paper relies on high-performance GPU servers, and the FGM algorithm has the issue of a linear assumption. Strategies such as model pruning can be employed, and the latest adversarial training methods can be explored, such as FreeLB (Free Large Batch Adversarial Training).

# Appendix

Table 8   Classification of 14 types of online fraud cases

| Case Classification |
| --- |
| Brushing and Rebate Fraud |
| Fake Shopping and Services |
| Other Types |
| Nude Chat Extortion |
| False Trading of Online Game Products |
| Impersonation of E-commerce Logistics Customer Service |
| Online Investment Platforms |
| False Credit Reporting |
| Impersonation of Leaders or Acquaintances |
| Online Dating and Socializing (Non-"Pig Slaughtering" Scheme) |
| Loans and Credit Card Processing Services |
| Impersonation of Public Security, Procuratorial, and Judicial Authorities, and Government Agencies |
| Pig Slaughtering Scheme |

# Acknowledgements

We would like to express our sincere gratitude to all those who have contributed to this research. First and foremost, we would like to thank the B City Public Security Bureau for providing the telecom fraud police data, which supported the completion of the experiments in this paper. Secondly, we appreciate the experimental environment provided by the People's Public Security University of China, which ensured the steady progress of the research. Additionally, we are also grateful to the Beijing Natural Science Foundation for their financial support. Finally, we extend our gratitude to all those who have contributed to this research, including my parents, teachers, classmates, boyfriend Shi Tianyang and laboratory colleagues. .

# Declarations of Conflict of interest

The authors declared that they have no conflicts of interest to this work.

# CRediT statements

**Liu Zhuoxian**: Conceptualization, Methodology, Software, Writing – Original Draft **Shi Tuo**: Data Curation, Project administration, Funding acquisition **Hu Xiaofeng**: Supervision, Validation